# Modern Türkiye'de Kişi Adları

**Personal Names in Modern Turkey**

Amaç Herdağdelen

amac@herdagdelen.com (Çilek Ağacı)

**Abstract**

We analyzed the most common 5000 male and 5000 female Turkish names based on their etymological, morphological, and semantic attributes. The name statistics are based on all Turkish citizens who were alive in 2014 and they cover 90% of all population. To the best of our knowledge, this study is the most comprehensive data-driven analysis of Turkish personal name inventory. Female names have a greater diversity than male names (e.g., top 15 male names cover 25% of the male population, whereas top 28 female names cover 25% of the female population). Despite their diversity, female names exhibit predictable patterns. For example, certain roots such as GÜL and NAR (*rose* and *pomegranate/red*, respectively) are used to generate hundreds of unique female names. Turkish personal names have their origins mainly in Arabic, followed by Turkish and Persian. We computed overall frequencies of names according to broad semantic themes that were identified in previous studies. We found that foreign-origin names such as OLGA and KHALED, pastoral names such as YAĞMUR and DENİZ (*rain* and *sea*, respectively), and names based on fruits and plants such as FİLİZ and MENEKŞE (*sprout* and *violet,* respectively) are more frequently observed among females. Among males, names based on animals such as ARSLAN and YUNUS (*lion* and *dolphin*, respectively) and names based on famous and/or historical figures such as MUSTAFA KEMAL and OĞUZ KAĞAN (founder of the Turkish Republic and the founder of the Turks in Turkish mythology, respectively) are observed more frequently.

**Öz**

2014 yılında hayatta olan Türkiye Cumhuriyeti vatandaşları arasında en sık gözlenen 5000'er erkek ve kadın ismini köken, biçim ve anlam bilimsel yönlerden inceledik. Bu çalışma, bilgimiz dahilinde Türkçedeki kişi adları istatistikleri üzerine yapılmış en kapsamlı çalışmadır. Kadın isimleri erkeklerinkinden daha çok çeşitlilik göstermektedir (örneğin en yaygın 15 erkek ismi her dört erkekten birisinde gözlenirken, kadınlarda nüfusun dörtte birini kapsamak için en yaygın 28 isim gerekmektedir). Kadın isimlerinin, çeşitliliklerine karşın, belirli örüntüler takip ettiveğini, GÜL ve NAR gibi az sayıdaki birkaç isim ile yüzlerce yeni isim türetildiğini gözledik. Türkiye'deki isimler ağırlıklı olarak Arapça, ardından Türkçe ve Farsça kökenlidir. Geçmiş çalışmalarda ele alınmış tematik listelere göre isimleri derledik ve sıklık dağılımlarını inceledik. Buna göre yabancı kökenli isimler (OLGA, KHALED), doğadan esinlenen isimler (YAĞMUR, DENİZ) ve meyve/bitki kökenli isimler (FİLİZ, MENEKŞE) kadınlar arasında daha sık gözlenmekteyken, hayvan isimleri (ARSLAN, YUNUS) ve ünlü ya da tarihi kişiliklerden esinlenen isimler (MUSTAFA KEMAL, OĞUZ KAĞAN) erkekler arasında daha sık gözlenmektedir.

## 1 Giriş

Kişi adları bireyi diğer insanlardan ayırt etmeye yarayan sembollerdir (Köse, 2014). Ancak bu işlevin yanı sıra kişi adları aynı zamanda ad koyanların, adı konan çocuk hakkındaki beklentileri, hayata bakışları, toplum içinde kendilerini gördükleri ya da görmek istedikleri sosyo-ekonomik konum, ailelerinin etnik ve dini arkaplanları gibi alanlarda işleyen göstergebilimsel işaretlerdir. Bu yönleriyle kişi adları, bireysel unsurlar olmaktan çıkıp toplumsal ve kültürel mirasın bir parçası haline gelirler (Karahan, 2008; Köse, 2014; Canatan, 2012; Olenyo, 2011).

Kişi adları toplum dinamiklerini incelemek isteyen araştırmacılar için de zengin bir bilgi kaynağıdır. Bireyin adının belirlenmesinde rol oynayan etkenler arasında, bireyi adlandıranların içinde bulunduğu sosyo-ekonomik koşullar ve kişilik özelliklerinin yanısıra, bireyin doğduğu yıllarda topluma etki eden isimlendirme alışkanlıklarını ve modaları, kişinin doğduğu yıllarda yaşanan toplumsal dönüşümleri ve yaşamış meşhur insanların isimlerini sayabiliriz (İlhan, 2013; Sakaoğlu, 2001; Canatan, 2012; Köse, 2014; Manu Magda, 2012; Acıpayamlı, 1992).

Kısacası kişi adları, bireysel, toplumsal, ekonomik ve kültürel seviyelerde işleyen farklı dinamiklerin ortak etkileşiminin bir sonucudur (Zengin, 2014). Topluluk seviyesindeki isim istatistikleri, yukarıdaki saydığımız etkenlerin hem yıllar içinde nasıl değiştiğini hem de kendi aralarında nasıl etkileştiklerini anlamak için bize yardımcı olabilir. Bu yönüyle kişi adları toplumsal dinamiklerdeki değişimin izini sürmemize yardımcı olabilecek fosil kayıtları gibidir.

## 2 Motivasyon

*Kişi adları bilimi* (İng. *anthroponomastics* ya da *anthroponymy*), dil biliminin insan adlarını inceleyen bir alt dalıdır. Bu alanla dil bilimcilerin yanı sıra toplum bilimi, antropoloji, tarih gibi çeşitli disiplinlerden araştırmacılar da ilgilenir (Manu Magda, 2012).

Türkiye ve Türk kültüründeki kişi adları hakkında geliştirilmiş geniş bir literatür bulunmaktadır (Acıpayamlı, 1992; Kalkandelen, 2008; Kibar, 2005). Kullandıkları teknik ve konuları bakımından mevcut çalışmaları şu ana başlıklar altında ele alabiliriz:

- İsimlendirme gelenekleri (Acıpayamlı, 1992; Sakaoğlu, 2001; Çelik, 2007; Kibar, 2005).
- Yerel araştırmalar (Doğan vd., 2013; Mehrali, 2014; Altun, 2011; Sarıtaş, 2009).
- Dilbilimsel araştırmalar (Duman, 2004).
- Tarihsel çalışmalar (ör. Osmanlı temettüat defterlerine geçen isimlerin tasnifi) (Hazar, 2008; Kurt, 1996; Kurt, 2005).
- Tematik çalışmalar (ör. Türk isimlendirme kültüründe renklerin, hayvanların, doğanın yeri) (Şenel vd., 2008; Boz, 2004) .
- Toplum bilimi açısından isimlendirme alışkanlıkları (ör. kitle iletişim araçlarının isimlendirme alışkanlıkları üstündeki etkisi) (Canatan, 2012; Çelik, 2007; Karahan, 2008; Köse, 2014; Uysal, 2012).

Bu çalışmaların bazılarında isimlendirme gelenekleri ve toplumun isimlendirme konusundaki düşünceleri tasnif, anket ve yüz yüze görüşme teknikleriyle incelenmiştir (Acıpayamlı, 1992; Kibar, 2005; Sarıtaş, 2009; Uysal, 2012). Duman (2004), anket yöntemiyle sınırlı sayıdaki katılımcılardan 1178 isimlik bir Türkçe kişi adları dökümü çıkarmış ve bu isimlerin köken, anlam ve biçim yönünden analizini yapmıştır.

Kültürel mirasın parçası olarak kişi adlarının seçildiği havuza etki eden dil bilimsel unsurlara bakılmış, tematik listeler çıkarılmış, toplu iletişim araçlarının isimlendirme alışkanlıkları üstündeki etkisine bakılıp, isimlerin göstergebilimsel açıdan nelere işaret ettiği tartışılmıştır (Zengin, 2014; Şenel vd., 2008; Çelik, 2007; Boz, 2004; Bayraktar, 2014; Köse, 2014). Yerel ve tarihi isim listelerini derleyen ve inceleyen çalışmalar veri analizinden faydalanan diğer örnekler arasındadır (Altun, 2011; Doğan vd., 2013; Hazar, 2008; Kurt, 1996; Kurt, 2005; Mehrali, 2014).

Yukarıda örneklendirdiğimiz çalışmalar çeşitli disiplinlerden araştırmacılar tarafından yürütülmekte ve hem kullanılan veri hem yöntem açısından büyük bir zenginlik göstermektedirler.

Bu çalışmaya başlarken amacımız modern Türkiye'deki isim istatistiklerinin saptamasını yapmak ve daha özel alanlara odaklanmış araştırmacıların ellerindeki veriyi karşılaştırabilecekleri, toplumun genelini temsil eden bir referans noktası oluşturmaktı. Çalışmamızı gerçekleştirirken, özel bir alana odaklanmak yerine geniş kapsamlı bir veri kümesi üzerinden kişi adlarının betimleyici bir analizini yapmayı hedefledik. Bu sayede kişi adları bilimi alanında çalışacak dil bilimcilerin ve diğer disiplinlerden araştırmacıların faydalanabilecekleri bir referans oluşturmayı ümit ediyoruz.

## 3 Veri

Kullandığımız veriyi Türkiye İstatistik Kurumu'ndan (TÜİK) bilgi edinme kanunu kapsamında yaptığımız başvuru ile elde ettik. Bu veri 2014 yılında hayatta olan tüm Türkiye Cumhuriyeti vatandaşları arasında en sık gözlenen 5000 erkek ve 5000 kadın isminin kullanım sıklıklarını içermekte olup, toplamda 38.7 milyon kadının (kadın nüfusunun yüzde 92'sinin) ve 39.0 milyon erkeğin (erkek nüfusunun yüzde 91'inin) isim istatistiklerini kapsamaktadır. Elimizdeki veri isimlerin toplam gözlenme sayılarından ibaret olup birey bazında ayırt edici bir bilgi içermemektedir.

*Tablo 1. Genel isim istatistikleri.*

| | Toplam Nüfus | Gözlenen Kişi | Tekil İsim | Tekil İlk İsim |
|---|---|---|---|---|
| Kadın | 38.711.602 | 35.462.143 (92%) | 5.000 | 3.767 |
| Erkek | 38.984.302 | 35.517.608 (91%) | 5.000 | 2.473 |

Çift isimler (göbek adı kullanılan kayıtlar) veride ayrı kayıtlar olarak yer almaktadır. Örneğin Türkiye'de toplam 784.731 kişi ALİ ve 152.061 kişi MEHMET ALİ ismindedir. Çift isimleri incelediğimiz bölümler haricinde isim sıklıklarını hesaplarken, çift isimli kayıtlarda yer alan birinci ismi göbek adı olarak kabul edip yok saydık. Örneğin ALİ isminin toplam gözlenme sıklığını hesaplarken ALİ, MEHMET ALİ, DURMUŞ ALİ gibi isimlerin sıklıklarını topladık.

Ancak MEHMET ALİ'lerin sayısı MEHMET isminin sıklığının hesaplanmasında kullanılmadı.

*Tablo 2. En yaygın ilk isimler.*

| Kadın ismi | Kişi sayısı | Oran (%) | Erkek ismi | Kişi sayısı | Oran (%) |
|---|---|---|---|---|---|
| FATMA | 1.280.005 | 3,31 | MEHMET | 1.421.301 | 3,65 |
| AYŞE | 991.553 | 2,56 | MUSTAFA | 1.162.547 | 2,98 |
| EMİNE | 847.879 | 2,19 | ALİ | 1.131.581 | 2,90 |
| HATİCE | 757.858 | 1,96 | AHMET | 932.010 | 2,39 |
| ZEYNEP | 613.293 | 1,58 | HÜSEYİN | 691.479 | 1,77 |
| NUR | 591.211 | 1,53 | İBRAHİM | 625.874 | 1,61 |
| ELİF | 480.302 | 1,24 | HASAN | 588.220 | 1,51 |
| MERYEM | 294.025 | 0,76 | MURAT | 550.772 | 1,41 |
| MERVE | 281.944 | 0,73 | YUSUF | 495.563 | 1,27 |
| ZEHRA | 266.405 | 0,69 | İSMAİL | 472.306 | 1,21 |
| ESRA | 243.164 | 0,63 | EMRE | 404.698 | 1,04 |
| ÖZLEM | 234.502 | 0,61 | ÖMER | 397.389 | 1,02 |
| SULTAN | 214.775 | 0,55 | OSMAN | 381.491 | 0,98 |
| YASEMİN | 207.768 | 0,54 | RAMAZAN | 377.320 | 0,97 |
| BÜŞRA | 204.027 | 0,53 | CAN | 321.048 | 0,82 |
| KÜBRA | 202.760 | 0,52 | HALİL | 298.049 | 0,76 |
| HÜLYA | 197.366 | 0,51 | FATİH | 288.979 | 0,74 |
| DİLEK | 190.239 | 0,49 | ABDULLAH | 276.123 | 0,71 |
| MELEK | 185.207 | 0,48 | SÜLEYMAN | 238.983 | 0,61 |
| SONGÜL | 177.366 | 0,46 | HAKAN | 231.655 | 0,59 |
| LEYLA | 176.732 | 0,46 | FURKAN | 218.395 | 0,56 |
| SEVİM | 174.765 | 0,45 | EMİN | 217.748 | 0,56 |
| FADİME | 171.854 | 0,44 | SALİH | 207.540 | 0,53 |
| AYSEL | 169.870 | 0,44 | KADİR | 205.488 | 0,53 |
| ŞERİFE | 169.507 | 0,44 | ENES | 203.903 | 0,52 |
| RABİA | 168.849 | 0,44 | KEMAL | 203.193 | 0,52 |
| HACER | 167.557 | 0,43 | BURAK | 200.050 | 0,51 |
| HAVVA | 161.343 | 0,42 | ADEM | 198.280 | 0,51 |
| TUĞBA | 158.809 | 0,41 | RECEP | 196.857 | 0,50 |
| HANİFE | 158.788 | 0,41 | MAHMUT | 195.703 | 0,50 |

## 4 Sonuçlar

Bulgularımız şöyle özetlenebilir.

- İnsanların büyük bir kısmı nüfus kaydında tek isme sahip. Çift isme sahip kadınlarda ilk isim sıklıkla NUR olarak tercih edilmiş. Erkek isimlerinde ise göze çarpan bir örüntü yok.
- Kadın isimleri erkek isimlerinden daha çeşitli. Benzer büyüklükteki kadın ve erkek nüfuslarını kapsayan örneklemde 3767 tekil kadın ismi ve 2473 tekil erkek ismi gözleniyor.
- Kadın isimlerindeki çeşitliliğin önemli bir kısmı NUR, GÜL, NAZ gibi diğer isimlerin başına veya sonuna getirilerek elde edilen yeni birleşik isimlerinden kaynaklanıyor.
- Etimolojik açıdan Türkçedeki isimlerin en yaygın kaynağı Arapça. Ardından sırasıyla Türkçe ve Farsça geliyor. Ayrıca erkek isimlerinde İbranice ve kadın isimlerinde Rumca/Yunanca kökenli isimlerin sayısı dikkat çekiyor.
- Orjinal haliyle ya da Türkçe ses yapısına uydurulmuş şekliyle kullanılan yabancı isimler (ör. OLGA, NATALİA, ANNA, vs.) özellikle kadınlar arasında daha sık gözleniyor.

- Bitki ya da meyve isimlerinin erkek nüfusunda gözlenme oranı yüzde 0,3. Kadın nüfusunda gözlenme oranı ise yüzde 1,3.
- Yabancı isimlerin erkek nüfusunda gözlenme oranı yüzde 0,01. Kadın nüfusunda gözlenme oranı ise yüzde 0,1.
- Doğadan esinlenen isimlerin erkek nüfusunda gözlenme oranı yüzde 0,7. Kadın nüfusunda gözlenme oranı ise yüzde 1,7.
- Hayvan isimlerinin erkek nüfusunda gözlenme oranı yüzde 0,8, kadın nüfusunda gözlenme oranı ise yüzde 0,3.
- Tarihi kişilikler, sanatçılar ve halen yaşayan veya ölmüş politikacıların isimlerinden esinlenilerek verilmiş isimler (ör. MUSTAFA KEMAL, OĞUZ KAĞAN, RECEP TAYYİP, YUSUF İSLAM, vs.) yaygınlıkları az olsa da isimlendirme kültüründe kendisine yer bulmuş ve özellikle erkekler için tercih edilen isimler olarak göze çarpıyor.

### *4.1 Çift ve Tek İsim Kullanımı*

*Şekil 1: Tek ve çift isimlerin cinsiyete göre dağılımı*

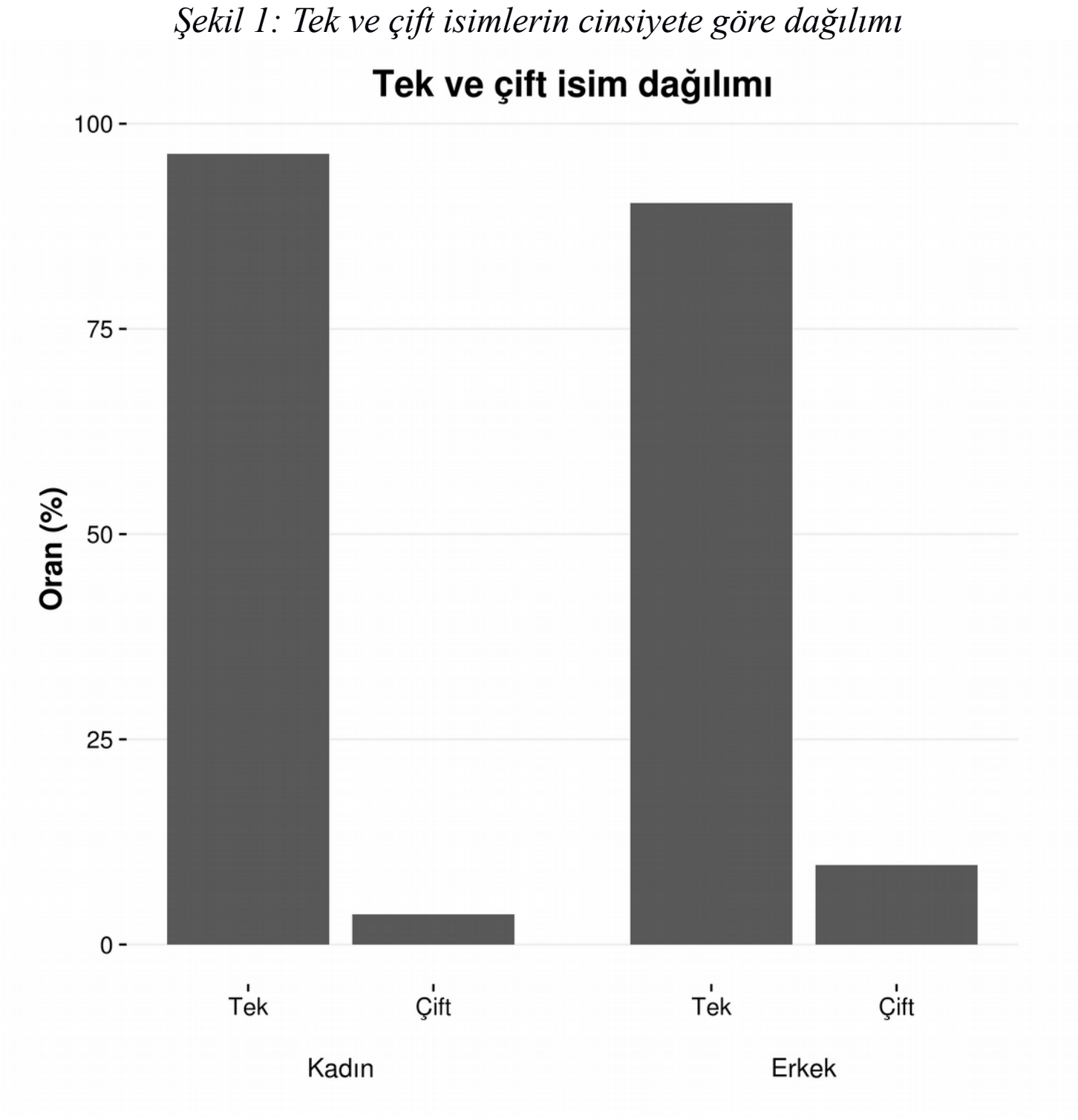

Toplam nüfus içerisinde her on erkekten birisi (%9,7) nüfusa çift isimle kayıtlı. Kadınlarda ise bu oran çok daha düşük (%3,7). Örneklemimizde kendine yer bulan tek üç isimli kayıt 429 erkeğin paylaştığı FATİH SULTAN MEHMET ismi.

### *4.2 En Yaygın Çift İsimler*

Kadınlar arasında NUR, çift isimlerde en çok tercih edilen ilk isim olarak çekiyor. En yaygın 15 çift isimden 10 tanesi NUR ismi kullanılarak oluşturulmuş. Erkeklerdeki çift isimler ise daha fazla çeşitlilik gösteriyor.

*Tablo 3. En yaygın çift isimler.*

| Kadın ismi | Kişi sayısı | % | Erkek ismi | Kişi sayısı | % |
|---|---|---|---|---|---|
| FATMA NUR | 40.222 | 0,11 | MEHMET ALİ | 152.061 | 0,43 |
| AYŞE NUR | 33.611 | 0,09 | YUNUS EMRE | 103.905 | 0,29 |
| NİSA NUR | 28.518 | 0,08 | HALİL İBRAHİM | 94.631 | 0,27 |
| BEYZA NUR | 26.199 | 0,07 | ÖMER FARUK | 94.415 | 0,27 |
| ELİF NUR | 23.470 | 0,07 | MEHMET EMİN | 78.677 | 0,22 |
| ESMA NUR | 21.593 | 0,06 | HASAN HÜSEYİN | 57.308 | 0,16 |
| AYŞE GÜL | 19.058 | 0,05 | İBRAHİM HALİL | 48.905 | 0,14 |
| HATİCE KÜBRA | 17.893 | 0,05 | MUHAMMED ALİ | 42.610 | 0,12 |
| FATMA ZEHRA | 17.066 | 0,05 | ALİ OSMAN | 36.661 | 0,10 |
| EDA NUR | 16.740 | 0,05 | MUHAMMED EMİN | 32.816 | 0,09 |

### *4.3 İsimlerde Çeşitlilik*

Analizin geri kalanında çift isimlerde ilk ismi göbek adı olarak kabul edip yok saydık. Örneğin MEHMET ALİ'lerin sayısı ALİ'lere eklendi. Kadın ve erkek isimlerindeki çeşitliliği karşılaştırırken ilk dikkat çeken nokta tekil kadın isimlerinin sayısının tekil erkek isimlerinin sayısından çok daha yüksek olması. Elimizdeki örneklemde temsil edilen erkek ve kadın sayısının birbirine çok yakın olmasına karşın 2474 tekil erkek ismine karşılık 3767 tekil kadın ismi gözledik.

En yaygın 15 erkek ismi toplam erkek nüfusunun %25'i tarafından paylaşılırken, en yaygın 28 kadın ismi kadınların %25'i tarafından paylaşılıyor. Benzer şekilde, erkek nüfusunun yarısını kapsamak için en yaygın 79 isim yeterliyken, kadın nüfusunun yarısı en yaygın 126 isim ile kapsanabiliyor.

*Şekil 2: Kadın ve erkek isimlerinin kümülatif gözlenme dağılımları. Yatay eksen isimlerin gözlenme sıklığına göre sıralamasını belirtmektedir. Dikey eksen karşılık gelen sıraya kadar olan en yaygın isimlerin nüfusun ne kadarı tarafından paylaşıldığını belirtir.*

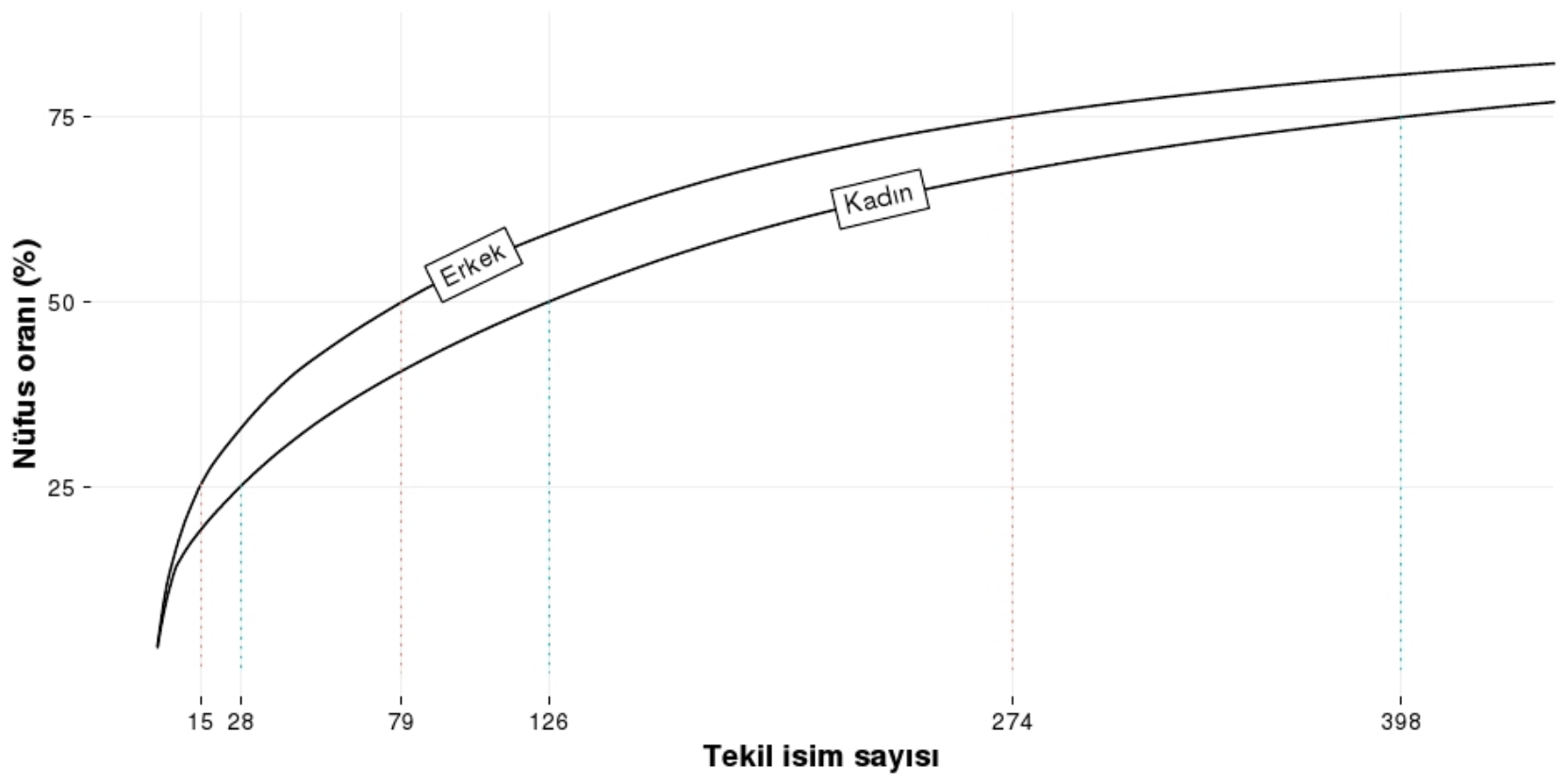

Kadın isimlerinde çeşitliliğin erkek isimlerindekinden daha çok olduğunun başka bir sayısal göstergesi rastgele seçilen iki erkeğin ve iki kadının aynı isme sahip olma olasılıkları. Elimizdeki veriye göre rastgele seçilen iki erkeğin aynı isme sahip olma olasılığı 120'de 1 iken (%0,83) iken rastgele seçilen iki kadının aynı isme sahip olma olasılığı 175'te 1 (%0,57).

Sıradaki şekilde en yaygın 20'şer erkek ve kadın isminin gözlenme sıklıklarını görselleştirdik. Görülebileceği gibi, genel nüfustaki erkek ve kadın sayıları birbirilerine yakın olmasına karşın, en yaygın isimler arasında erkek isimleri sıralamada denk gelen kadın isimlerinden çok daha fazla gözlenme sıklığına sahip. Bu da kadın isimlerinin daha çok çeşitlilik gösterdiği gözlemimizle uyumlu.

*Şekil 3: En yaygın kadın ver erkek isimlerinin gözlenme sıklıklarındaki değişim. Yatay eksen ismin gözlenme sırasını belirtmektedir. Dikey eksen karşılık gelen sıradaki ismin nüfusta kaç defa gözlendiğini belirtir.*

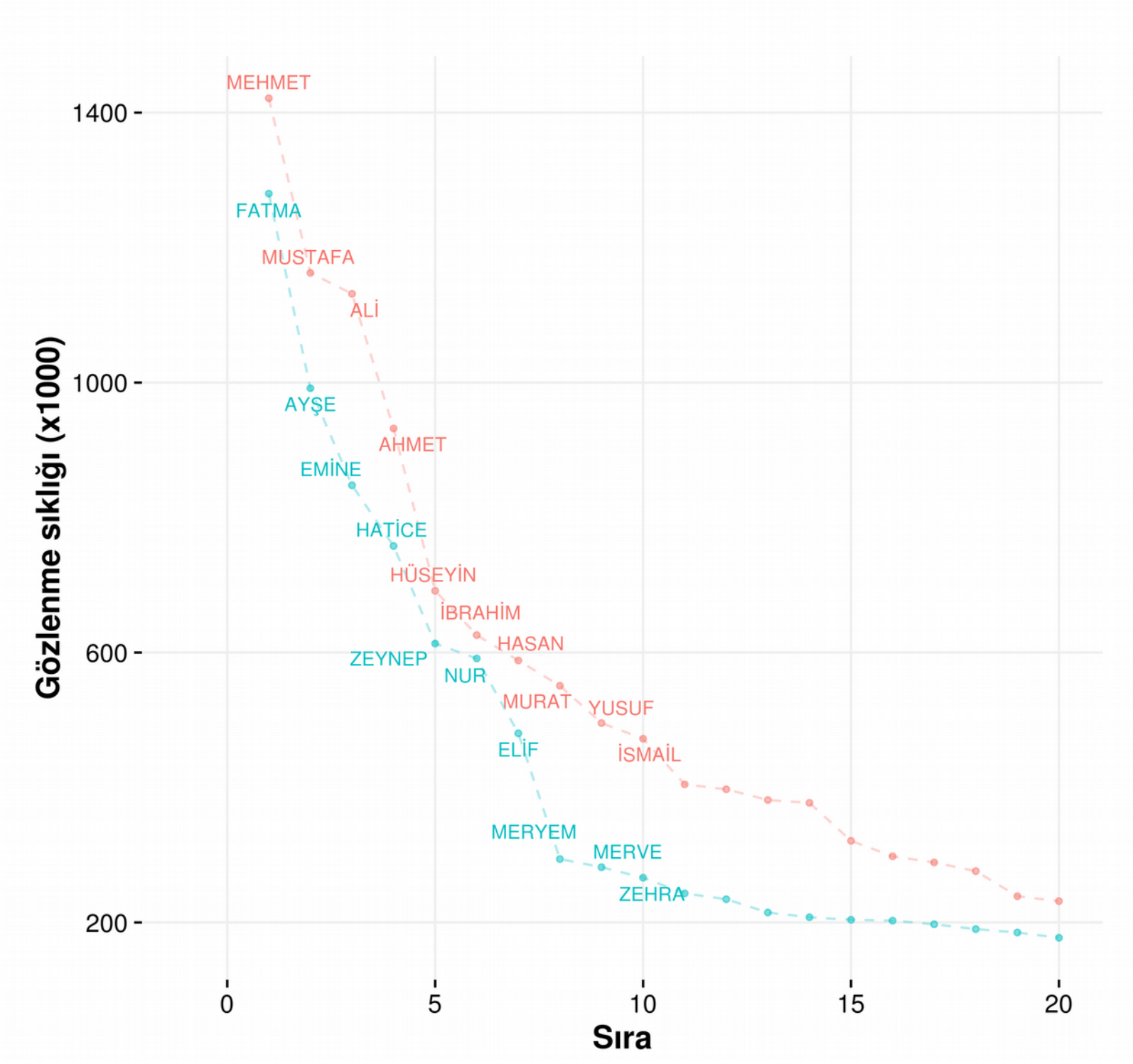

*4.3.1 İsim kökleri ve ekleri*

Daha önce belirttiğimiz gibi kadın isimlerinin sayısal olarak erkek isimlerine göre daha fazla çeşitlilik gösterdiğini gözledik. Ancak veriye baktığımızda kadın isimlerindeki çeşitliliğin belli bazı örüntüler takip ettiğini ve isim köklerinin daha az çeşitlilik gösterdiğini görüyoruz. Örneğin veride NUR kökü ile başlayan 67 farklı kadın ismi gözledik (NURCAN, NURAY, NURTEN, NURİYE, NURAN bunların en yaygın gözüken beş örneği). Benzer şekilde NUR ile biten tam 188 farklı kadın ismi gözledik (AYNUR, İLKNUR, AYŞENUR, ÖZNUR, NİSANUR bunların en yaygın gözüken beş örneği). Kadın isimlerinde sık rastlanan diğer isim kökleri arasında GÜL, NAZ ve SU gözleniyor. Erkek ve kadın isimlerini bu boyutta karşılaştırmak için her iki isim listesinde diğer isimlerin başında ve sonunda en sık gözlenen (ve birbirlerini kapsamayan) isimleri belirledik. Örneğin ABDUL verimizde gözlenen bir erkek ismi olduğu için ve "A", "Ab", "Abd" veya "Abdu" bir isim olarak veride gözlenmediği için ABDUL ile başlayan tüm erkek isimlerinin sayısını hesapladık bunu ABDUL'e bağladık. İsimlerin başında yer alan en yaygın 10 isim köküne baktığımızda aşağıdaki listeyi elde ediyoruz. Bu yöntemle ALİ ve ALİM ya da SU ve SUPHİYE gibi aralarında köken ya da biçim yönünden bir ilişki olmayan ikililer de listelense bile oldukça düşük bir hata oranı ile tüm isimleri otomatik olarak işleme imkanı bulduk. Burada verdiğimiz istatistikleri kesin sonuçlar yerine isim biçimlerini incelemek için kullanılabilecek yol göstericiler olarak dikkate almak daha doğru olur.

*Tablo 4. En sık gözlenen kadın ismi kökleri. Her kök için o kökle başlayan isim sayısı ve en yaygın 10 isim verilmiştir.*

| Kök | İsim | Örnekler |
|---|---|---|
| GÜL | 191 | GÜLSÜM, GÜLAY, GÜL, GÜLER, GÜLCAN, GÜLTEN, GÜLŞEN, GÜLLÜ, GÜLSEREN, GÜLBAHAR |
| NUR | 67 | NUR, NURCAN, NURAY, NURTEN, NURİYE, NURAN, NURGÜL, NURSEL, NURHAN, NURDAN |
| SU | 40 | SULTAN, SUDE, SUNA, SU, SUZAN, SUDENAZ, SUDENUR, SUNAY, SUPHİYE, SUEDA |
| NAZ | 35 | NAZ, NAZLI, NAZMİYE, NAZİIFE, NAZAN, NAZİRE, NAZLICAN, NAZİME, NAZİK, NAZAR |
| CAN | 15 | CANAN, CANSU, CANSEL, CANDAN, CAN, CANSEVER, CANGÜL, CANSER, CANSIN, CANPAŞA |
| AYŞE | 14 | AYŞE, AYŞEGÜL, AYŞENUR, AYŞEN, AYŞENAZ, AYŞETE, AYŞEDUDU, AYŞEHAN, AYŞEANA, AYŞELİ |
| SEVİ | 14 | SEVİM, SEVİL, SEVİNÇ, SEVİLAY, SEVİYE, SEVİN, SEVİLCAN, SEVİ, SEVİLE, SEVİMNUR |
| ÜMMÜ | 14 | ÜMMÜGÜLSÜM, ÜMMÜ, ÜMMÜHAN, ÜMMÜGÜL, ÜMMÜHANİ, ÜMMÜGÜLSÜN, ÜMMÜYE, ÜMMÜŞ, ÜMMÜŞEN, ÜMMÜHANI |
| ŞEN | 12 | ŞENGÜL, ŞENAY, ŞENNUR, ŞENEL, ŞENİZ, ŞENER, ŞENOL, ŞENCAN, ŞENGÜN, ŞENAL |
| LAL | 10 | LALE, LALİHAN, LALİZER, LAL, LALEŞ, LALİZAR, LALA, LALI, LALEHAN, LALEZAR |

En üretken 10 kadın ismi kökü 412 isme kaynaklık ederken, erkeklerde bu sayı 178 oldu. En üretken 10 kadın ismi son ek olarak 577 ismin sonunda yer alırken, erkeklerde bu sayı 304'tür.

*Tablo 5. En sık gözlenen erkek ismi kökleri. Her kök için o kökle başlayan isim sayısı ve en yaygın 10 isim verilmiştir.*

| Kök | İsim | Örnekler |
|---|---|---|
| ABDUL | 51 | ABDULLAH, ABDULKADİR, ABDULSAMET, ABDULKERİM, ABDULBAKİ, ABDULAZİZ, ABDULVAHAP, ABDULHAMİT, ABDULHAKİM, ABDULSELAM |
| NUR | 23 | NURİ, NURETTİN, NURULLAH, NUR, NUREDDİN, NURİTTİN, NURHAN, NURDOĞAN, NURAL, NURİCAN |
| ALİ | 19 | ALİ, ALİCAN, ALİHAN, ALİM, ALİŞAN, ALİRIZA, ALİOSMAN, ALİSEYDİ, ALİEKBER, ALİIHSAN |
| ABDÜL | 18 | ABDÜLKADİR, ABDÜLSAMET, ABDÜLKERİM, ABDÜLBAKİ, ABDÜLAZİZ, ABDÜLHAMİT, ABDÜLSAMED, ABDÜLMECİT, ABDÜLSELAM, ABDÜLLATİF |
| CAN | 15 | CAN, CANER, CANBERK, CANİP, CANPOLAT, CANAN, CANDAN, CANTÜRK, CANKAT, CANDAŞ |
| BERK | 13 | BERKAY, BERK, BERKE, BERKAN, BERKANT, BERKCAN, BERKİN, BERKER, BERKEN, BERKEHAN |
| ŞAH | 11 | ŞAHİN, ŞAH, ŞAHABETTİN, ŞAHAN, ŞAHAP, ŞAHİSMAİL, ŞAHMETTİN, ŞAHBETTİN, ŞAHMURAT, ŞAHABEDDİN |
| ALP | 10 | ALPER, ALPEREN, ALP, ALPARSLAN, ALPASLAN, ALPAY, ALPTEKİN, ALPTUĞ, ALPHAN, ALPCAN |
| ATA | 10 | ATAKAN, ATA, ATALAY, ATABERK, ATAHAN, ATACAN, ATANUR, ATAY, ATAMAN, ATABEY |
| CEM | 8 | CEMAL, CEMİL, CEM, CEMALETTİN, CEMALİ, CEMRE, CEMALEDDİN, CEMAİL |

## *4.4 Köken bilimsel analiz*

Türkçedeki kişi adlarına kaynaklık eden dillerin dağılımını incelemek için elimizdeki kişi adlarını Türk Dil Kurumu'nun İnternet sitesinde yayınladığı kişi adları sözlüğüyle eşleştirdik. Bu sözlükte kişi adlarının anlamlarının yanı sıra etimolojik köken bilgisi de yer almaktadır. Sözlükte yer alan toplam 1540 erkek isminin (örneklemdeki erkeklerin %97'si) ve 1709 kadın isminin (örneklemdeki kadınların %92'si) kökenlerinin dağılımına baktık.

Buna göre Türkçedeki yaygın isimlere en çok kaynaklık eden diller sırasıyla Arapça (766 erkek ve 733 kadın ismi), ardından Türkçe (551 erkek ve 388 kadın ismi) ve Farsça (92 erkek ve 272 kadın ismi). Bu üç dilin yanısıra

İbranice, Rumca, Moğolca da isimlere kaynaklık eden diğer dillerden. Burada göze çarpan bir eksiklik Kürtçe. Ancak Türk Dil Kurumu'nun çevrimiçi sözlüğünde belirtilen diller arasında Kürtçe yer almadığı için Kürtçe kökenli isimlerin oranı hakkında bir tahmin yapamıyoruz. Anekdotal olarak, Kürtçe kökenli olduğunu düşündüğümüz pek çok ismin sözlükte Farsça kökenli olarak belirtildiğini gözledik.

*Tablo 6. Kadın isimlerinin köken dil dağılımları ve her dil için en yaygın gözlenen üçer örnek isim.*

| Kaynak dil | İsim | İsim (%) | Kişi | Kişi (%) | Örnekler |
|---|---|---|---|---|---|
| Arapça | 733 | 42,9 | 19.705.790 | 60,2 | FATMA, AYŞE, EMİNE |
| Türkçe | 388 | 22,7 | 5.049.362 | 15,4 | ÖZLEM, DİLEK, SEVİM |
| Farsça | 272 | 15,9 | 3.904.391 | 11,9 | YASEMİN, DERYA, EBRU |
| Türkçe/Farsça | 135 | 7,9 | 1.629.913 | 5,0 | SONGÜL, AYTEN, TÜRKAN |
| Türkçe/Arapça | 79 | 4,6 | 802.588 | 2,5 | AYNUR, NURAY, İLKNUR |
| Farsça/Arapça | 65 | 3,8 | 656.975 | 2,0 | AYŞEGÜL, NURCAN, NURTEN |
| Rumca | 16 | 0,9 | 485.083 | 1,5 | FİLİZ, MELİSA, ELMAS |
| Moğolca | 5 | 0,3 | 214.855 | 0,7 | MERAL, CEREN, CEYLAN |
| Fransızca | 5 | 0,3 | 46.388 | 0,1 | BUKET, MÜGE, MANOLYA |
| Farsça/Rumca | 3 | 0,2 | 11.405 | <0,1 | GÜLFİDAN, GÜLDEMET, GÜLFİDE |
| İbranice | 2 | 0,1 | 6.794 | <0,1 | CEYHAN, TEVRAT |
| Türkçe/Fransızca | 1 | 0,1 | 77.657 | 0,2 | TÜLAY |

*Tablo 7. Erkek isimlerinin köken dil dağılımları ve her dil için en yaygın gözlenen üçer örnek isim.*

| Kaynak dil | İsim sayısı | İsim (%) | Kişi sayısı | Kişi (%) | Örnekler |
|---|---|---|---|---|---|
| Arapça | 766 | 49,7 | 21.612.473 | 62,9 | MEHMET, MUSTAFA, ALİ |
| Türkçe | 551 | 35,8 | 7.483.473 | 21,8 | EMRE, EFE, EREN |
| Farsça | 92 | 6,0 | 1.842.046 | 5,4 | CAN, HAKAN, MERT |
| Türkçe/Farsça | 55 | 3,6 | 448.397 | 1,3 | SERKAN, ERCAN, CANER |
| Türkçe/Arapça | 25 | 1,6 | 195.446 | 0,6 | EMIRHAN, BEDIRHAN, GÜRSEL |
| Farsça/Arapça | 17 | 1,1 | 142.309 | 0,4 | SERHAT, ALİHAN, ABUZER |
| İbranice | 16 | 1,0 | 2.379.935 | 6,9 | İBRAHİM, YUSUF, İSMAİL |
| Rumca | 6 | 0,4 | 38.819 | 0,1 | POYRAZ, TEMEL, MENDERES |
| Moğolca | 6 | 0,4 | 37.736 | 0,1 | OLCAY, KUBILAY, ERGUN |
| Fransızca | 2 | 0,1 | 70.709 | 0,2 | VOLKAN, SONAT |
| İngilizce | 1 | 0,1 | 36.081 | 0,1 | TAYFUN |

Yukarıda verdiğimiz dil dağılımının, kişi adlarının yaygınlığına göre nasıl değiştiğini gözlemek için adları gözlenme sıklıklarını kullanarak beş tane %20'lik dilime ayırdık. En tepedeki %20'lik dilim sırasıyla erkek ve kadın isimleri arasında en sık gözlenen isimleri içerirken, en alttaki dilim de elimizdeki veride en az gözlenen isimleri içermekte.

*Şekil 4: Kadın ve erkek isimlerinin gözlenme sıklıklarına göre belirlenmiş %20'lik dilimlerinin köken dil dağılımları. İsimleri nüfusta gözlenme sıklıklarına göre sıralayıp, en yaygın %20, ardından gelen %20 şeklinde beşer gruba böldük.*

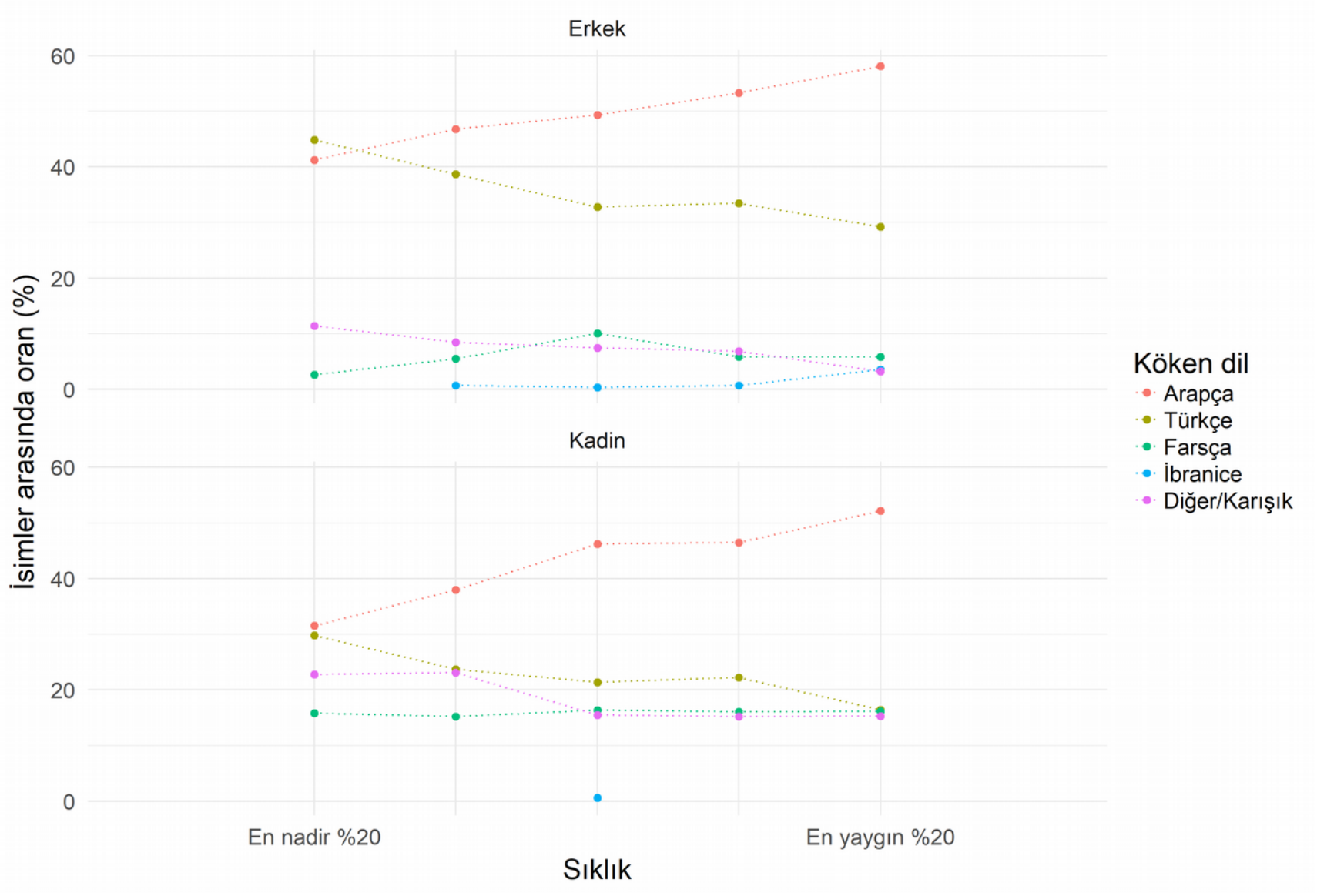

Buna göre Arapça kökenli adlar sık gözlenen isimler içinde daha yaygınken, Türkçe ve karışık kökenli adların oranı isimlerin gözlenme sıklığı azaldıkça artmakta.

## *4.5 Tematik İsim Listeleri*

### *4.5.1 Bitki ve meyveler*

Çeşitli kaynaklardan derlediğimiz (Boz, 2004; Zengin, 2014) bitki ve meyve isimlerini elimizdeki veriyle eşleştirdik. Bitki ve meyvelerden esinlenen isimlerin ağırlıklı olarak kadın ismi olarak kullanıldığını gözledik. Bitki ve meyvelerden esinlenmiş 9 tekil erkek ismi gözlenirken bu sayı kadınlarda 39 olarak gözlendi. Genel nüfusta her 100.000 erkekten 337 tanesi ve her 100.000 kadından 1256 tanesi bitki ve meyvelerden esinlenmiş bir isim taşımakta.

Tablo 8. Bitki ve meyve kökenli en yaygın kişi adları.

| Kadın | | | Erkek | | |
|---|---|---|---|---|---|
| İsim | Sıklık | Oran (her yüz binde) | İsim | Sıklık | Oran (her yüz binde) |
| SEMRA | 120.276 | 310,7 | BARAN | 61.615 | 158,1 |
| MELİKE | 111.986 | 289,3 | İDRİS | 47.042 | 120,7 |
| SÜMEYRA | 29.573 | 76,4 | BERKE | 19.250 | 49,4 |
| ÇAĞLA | 24.678 | 63,7 | GÖKÇE | 933 | 2,4 |
| GÖKÇE | 22.76 | 58,8 | ÇAKIR | 639 | 1,6 |
| DEFNE | 22.357 | 57,8 | SAMIR | 569 | 1.50 |
| BAŞAK | 21.216 | 54,8 | ÇAVUŞ | 493 | 1.30 |
| ÇİÇEK | 21.145 | 54,6 | BOSTAN | 376 | 1.00 |
| KARDELEN | 20.070 | 51,8 | ARGUN | 272 | 0.70 |
| MENEKŞE | 19.778 | 51,1 | | | |

### *4.5.2 Yabancı köken*

Bu bölümde, yabancı kökenli isimlerden kastımız başka dillerden Türkçeye geçmiş isimler değil ancak henüz modern Türkçede kendisine yer bulmamış isimleri inceledik. Verideki isimlerin üstünden geçerek yabancı kökenli isimleri işaretledik. Çıkan sonuçlarda, yabancı kökenli isimlerin çok büyük oranda kadınlarda gözlendiğini tespit ettik. Yabancı kökenli sadece 5 erkek ismi gözlenirken 55 tane yabancı kökenli kadın ismi gözlendi. Her 100.000 erkekten 8 tanesi ve her 100.000 kadından 81 tanesi yabancı kökenli bir isim taşımakta.

Yabancı erkek ve kadın isimleri arasındaki fark Türk vatandaşlığına evlilik yoluyla daha çok kadınların geçtiği bulgusuyla uyum içindedir (Altun ve Dinç, 2016).

*Tablo 9. Yabancı kökenli en yaygın kişi adları.*

| Kadın | | | Erkek | | |
|---|---|---|---|---|---|
| İsim | Sıklık | Oran (her yüz binde) | İsim | Sıklık | Oran (her yüz binde) |
| OLGA | 2.349 | 6,1 | MOHAMAD | 1.696 | 4,4 |
| HELEN | 1.911 | 4,9 | ELYESA | 474 | 1,2 |
| ANNA | 1.656 | 4,3 | KHALED | 394 | 1,0 |
| SVETLANA | 1.390 | 3,6 | MOUSTAFA | 317 | 0,8 |
| NATALIA | 1.322 | 3,4 | HOSSEIN | 274 | 0,7 |
| TATIANA | 1.128 | 2,9 | | | |
| MARYAM | 1.076 | 2,8 | | | |
| IRINA | 1.069 | 2,8 | | | |
| MARIA | 1.035 | 2,7 | | | |
| MARINA | 933 | 2,4 | | | |

### *4.5.3 Doğa*

Türk kültüründe isim koyarken kullanılan unsurlardan birisi de doğadır (Zengin, 2014; Kibar, 2005; Şenel vd al., 2008; Acıpayamlı, 1992). Verideki isimleri tek tek inceleyerek ve önceki çalışmalarda derlenmiş (Kibar, 2005) isimleri de ekleyerek, doğal öğelerden ve olaylardan esinlendiğini düşündüğümüz isimleri derledik. Buna göre her 100.000 erkekten 653 tanesi ve her 100.000 kadından 1744 tanesi doğa kökenli bir isim taşımakta. Gözlenen tekil isim sayıları ise erkekler ve kadınlar için sırasıyla 32 ve 27 oldu.

*Tablo 10. Doğal öğe ve olay kökenli en yaygın kişi adları.*

| Kadın | | | Erkek | | |
|---|---|---|---|---|---|
| İsim | Sıklık | Oran (her yüz binde) | İsim | Sıklık | Oran (her yüz binde) |
| YAĞMUR | 130.910 | 338,2 | DENİZ | 76.262 | 195,6 |
| BAHAR | 101.540 | 262,3 | EGE | 47.143 | 120,9 |
| DENİZ | 81.768 | 211,2 | AYAZ | 14.241 | 36,5 |
| DAMLA | 63.091 | 163,0 | POYRAZ | 13.443 | 34,5 |
| SU | 48.974 | 126,5 | DORUK | 13.410 | 34,4 |
| NEHİR | 46.397 | 119,9 | RÜZGAR | 13.320 | 34,2 |
| FİDAN | 36.651 | 94,7 | ŞAFAK | 11.670 | 29,9 |
| IRMAK | 34.512 | 89,2 | KUZEY | 9.850 | 25,3 |
| GÜNEŞ | 21.923 | 56,6 | KAYA | 9.091 | 23,3 |
| ALEV | 18.683 | 48,3 | DEMİR | 8.614 | 22,1 |

### *4.5.4 Hayvanlar*

Hayvan isimleri Türkçedeki kişi adlarına ilham veren başka bir kaynaktır (Şenel vd., 2008; Kibar, 2005). Biz elimizdeki veriyi tarayarak ve diğer kaynaklarda isim olarak kullanıldığı raporlanan hayvan isimlerini isim listemizle eşleştirerek bir liste derledik. Buna göre elimizdeki veride 10 erkek ve 11 kadın ismi hayvanlardan esinlenmiş. Her 100.000 erkekten 809 tanesi ve 100.000 kadından 341 tanesi hayvan isimlerinden esinlenmiş bir isim taşımakta.

*Tablo 11. Hayvanlardan esinlenen isimler.*

| Kadın | | | Erkek | | |
|---|---|---|---|---|---|
| İsim | Sıklık | Oran (her yüz binde) | İsim | Sıklık | Oran (her yüz binde) |
| CEREN | 77.367 | 199,9 | YUNUS | 130.420 | 334.5 |
| CEYLAN | 43.766 | 113,1 | ŞAHİN | 81.267 | 208,5 |
| KUMRU | 3.515 | 9,1 | DOĞAN | 65.771 | 168,7 |
| BÜLBÜL | 1.301 | 3,4 | ASLAN | 16.369 | 42,0 |
| GÜVERCİN | 1.122 | 2,9 | ARSLAN | 12.579 | 32,3 |
| GÖGERCİN | 1.115 | 2,9 | EJDER | 4.557 | 11,7 |
| TURNA | 1.043 | 2,7 | CEYLAN | 2.263 | 5,8 |
| KEKLİK | 1.040 | 2,7 | KARTAL | 919 | 2,4 |
| MARAL | 718 | 1,9 | KAPLAN | 689 | 1,8 |
| ÖRDEK | 656 | 1,7 | YUNÜS | 571 | 1,5 |

*4.5.5 Ünlüler*

Çalışmamızı gerçekleştiriken dikkatimizi çeken başka bir unsur erkek isimlerinde tarihi kişiliklerden ve yaşayan ya da ölmüş meşhur politikacı ve sanatçılardan esinlenmiş isimlerin varlığı oldu. Kadın isimlerinde benzer örneklere rastlamadık.

*Tablo 12. Tarihi, ünlü, siyasi kişiliklerden esinlenen erkek isimleri.*

| İsim | Kişi sayısı | Oran (her yüz binde) |
|---|---|---|
| MUSTAFA KEMAL | 16.964 | 43,5 |
| TAYYİP | 7.677 | 19,7 |
| YUSUF İSLAM | 6.945 | 17,8 |
| FATİH MEHMET | 6.927 | 17,8 |
| ECEVİT | 3.708 | 9,5 |
| NAMIK KEMAL | 2.963 | 7,6 |
| RECEP TAYYİP | 2.734 | 7,0 |
| ÖZAL | 2.098 | 5,4 |
| OĞUZ KAĞAN | 1.993 | 5,1 |
| YAHYA KEMAL | 1.197 | 3,1 |
| ÜMİT YAŞAR | 890 | 2,3 |
| OSMAN GAZİ | 638 | 1,6 |
| FATİH SULTAN | 637 | 1,6 |
| ŞAH İSMAİL | 622 | 1,6 |
| BİLAL HABEŞ | 611 | 1,6 |
| BİLGE KAAN | 601 | 1,5 |
| İLKER YASİN | 597 | 1,5 |
| MUHAMMET İKBAL | 559 | 1,4 |
| BATTAL GAZİ | 487 | 1,2 |
| YAŞAR KEMAL | 472 | 1,2 |
| SALTUK BUĞRA | 447 | 1,1 |
| FATİH SULTAN MEHMET | 429 | 1,1 |
| DEMİREL | 400 | 1,0 |
| ÖMER ŞERİF | 390 | 1,0 |
| ÖMER SEYFETTİN | 380 | 1,0 |
| KÖROĞLU | 371 | 1,0 |
| ORHAN VELİ | 357 | 0,9 |
| EYÜP SULTAN | 310 | 0,8 |
| MELİH ŞAH | 307 | 0,8 |
| TAYYİP ERDOĞAN | 303 | 0,8 |
| BÜLENT ECEVİT | 297 | 0,8 |
| MUSTAFA ATA | 294 | 0,8 |
| RECEP TAYİP | 285 | 0,7 |
| YILDIRIM BEYAZIT | 274 | 0,7 |
| SEZAR | 266 | 0,7 |
| GENÇ OSMAN | 254 | 0,7 |

## 5 Sonuç

Bu çalışmada Türkiye İstatistik Kurumu verilerine dayanarak 2014 itibariyle Türkiye'de kullanımda olan isimlerin betimleyici istatistiklerini ve çeşitli temalardaki isimlerin dağılımlarını inceledik. Çalışmamız modern Türkçedeki isimlendirme alışkanlıklarının bir resmini çektiği gibi sonuçlarımızın ileride bu alanda yapılacak benzer çalışmalar için bir referans noktası olacağını ve Türkçe ad bilimi çalışmalarında veri biliminin katkılarına bir örnek oluşturacağını ümit ediyoruz.

İncelememizde yöntem ve veriden kaynaklanan çeşitli sınırlamalara da dikkat çekmek isteriz. Bu çalışmada 2014 yılında hayatta olan vatandaşların isimleri kullanıldığı için isimlendirme alışkanlıklarının yıldan yıla değişimlerine bakmamız mümkün olmadı. Doğum yılı ya da doğum yeri kırılımlı isim istatistikleri farklı yaş kesitlerindeki isim dağılımı farklılıklarını incelememize olanak sağlayabilirdi. Aynı zamanda veri odaklı bir ad bilimi araştırmasına örnek oluşturmak istediğimiz için isimlerin göstergebilimsel ya da toplum bilimsel yönlerinin incelenmesi bu çalışmanın odağında yer almadı.

## Kaynaklar

Acıpayamlı, O. (1992). *Türk Kültüründe Ad Koyma Folkloru'nun Morfolojik ve Fonksiyonel Yönlerden İncelenmesi.* IV. Milletlerarası Türk Halk Kültürü Kongresi Bildirileri, IV. Cilt, Gelenek, Görenek ve İnançlar, Ofset Repromat Matbaası, Ankara.

Altun, H. O. (2011). *1975-2009 Yılları Arasında Bitlis'te Çocuklara Verilen Adlar.* Atatürk Üniversitesi Türkiyat Araştırmaları Enstitüsü Dergisi(45).

Altun, N. & Dinç A. (2016). *Yabancı Gelinlerin Türk Ailesi İçindeki Yerine Sosyolojik Bir Bakış.* Halk Kültüründe Aile Uluslararası Sempozyumu.

Bayraktar, N. (2014). *Türkçede Renk Adlarıyla Özel Ad Yapımı.* Journal Of Language and Linguistic Studies 9(2). 95--114.

Boz, E. (2004). *Türkiye Türkçesinde Kişi Adı Olarak Meyve Adları*. Afyon Kocatepe Üniversitesi Sosyalbilimler Enstitüsü Dergisi 7 (1), 1--13.

Canatan, K. (2012). *Türkiye'nin İsim Haritasının Temeli Olarak Ehl-İ Beyt Sevgisi.* Türk Kültürü ve Hacı Bektaş Veli Araştırma Dergisi(62).

Cooper, D. L. (2011). *From Karen To Keisha: The New Black Names.* Tyca-Southeast 44(1).

Çelik, C. (2007). *Bir Kimlik Beyanı Olarak İsimler: Kişi İsimlerine Sosyolojik Bir Yaklaşım.* Sosyoloji Araştırmaları Dergisi 2, 5–21.

Doğan, L. & Adalı Osman, D. (2013). *Yunanistan'ın Batı Trakya Bölgesi'ndeki Kozlukebir ve Köylerinde Kişi İsimleri.* Uluslararası Türk Dili ve Edebiyatı Kongresi (21).

Duman, D. (2004). *A Characterization Of Turkish Personal Name Inventory.* International Journal Of The Sociology Of Language, 155--178.

Hazar, M. (2008). *Artukogulları Zamanında Dedekorkut Kitabı'ndaki Kişi Adları.* Electronic Turkish Studies 3(1).

Kalkandelen, H. (2008). *Türkçemizde Kullanılan Kişi Adları.* Atatürk Üniversitesi Sosyal Bilimler Enstitüsü Dergisi 12(2).

İlhan, U. (2013), *Türkiye Türkçesinde Hayvan Adlarından Türetilmiş Bitki Adları.* Uluslararası Türkçe Edebiyat Kültür Egitim (TEKE) Dergisi 2(1).

Karahan, L. (2008). *Türkçede Dini Anlamlı Bazı Kişi Adlarını Ekle Degiştirme Geleneği,* In The 51st Meeting Of The Permanent International Altaistic Conference'.

Kibar, O. (2005). *Türk Kültüründe Ad Verme: Kişi Adları Üstüne Bir Tasnif Denemesi.* Vol. 194, Akçağ.

Kurt, Y. (2005). 1525 *Tarihinde Adana Sancagında Türkçe Kişi Adları Üzerine*. Bal-Tam Türklük Bilgisi 2, 72--83.

Kurt, Y. (1996). *Osmanlı Tahrir Defterlerinin Onomastik Degerlendirilmesine Uygulanacak Metod*. Osmanlı Araştırmaları 16(16).

Köse, A. (2014). *Degişimin Gölgesindeki Gelenek: Popüler Diziler ve Farklılaşan Ad Verme Kültürü*. Milli Folklor 26(101).

Manu Magda, M. (2012). P*ragmatics and Anthroponymy: Theoretical Considerations On The System Of Appellatives In Contemporary Romanian.* In Name and Naming: Synchronic and Diachronic Perspectives, O. Felecan (ed.), 18--31.

Mehrali, C. (2014). *Kişi Adları Üzerine Dilbilimsel Bir Çalışma (Ağrı İli Örnegi)*. Atatürk Üniversitesi Türkiyat Araştırmaları Enstitüsü Dergisi(52).

Olenyo, M. (2011). *What's In A Name? An Analysis Of The Semantics Of Lulogooli Personal Names*. International Journal Of Humanities and Social Science 1(20), 211--218.

Sakaoglu, S. (2001). *Türk Ad Bilimi*, Vol. 1, Türk Dil Kurumu.

Sarıtaş, S. (2009). *Balıkesir Üniversitesi Ögrencilerinin Günümüzdeki Adlar ve Ad Verme Hakkındaki Görüşleri.* Balikesir University Journal Of Social Sciences Institute 12(21).

Şenel, M. & İlhan, N. (2008). *Av Hayvanlarının Kişi Adlarındaki Yansıması.* Av ve Avcılık Kitabı, 321--336.

Uysal, B. (2012). *Alevi İnanç Sisteminde Adlar ve Dil-Kimlik İlişkisi*. Türk Kültürü ve Hacı Bektaş Veli Araştırma Dergisi(62).

Zengin, D. (2014). *Türk Toplumunda Adlar ve Soyadları. Sosyo-Kültürel ve Dilbilimsel Bir Yaklaşım.* Kurmay Yayınları.